\documentclass[letterpaper, 10 pt, conference]{ieeeconf}  

\usepackage{cite}
\usepackage{graphicx}
\usepackage{multicol}
\usepackage{amsfonts}
\usepackage{amsmath}
\usepackage{booktabs}
\usepackage{bbm}

\IEEEoverridecommandlockouts                              

\usepackage{newfloat}
\DeclareFloatingEnvironment[
    name={Table S\hspace{-0.28em}},
    fileext=lst,
    listname={List of Supplementary Tables}
]{supptable}

\DeclareFloatingEnvironment[name={Figure S\hspace{-0.28em}},fileext=lsf,listname={List of Supplementary Figures}]{suppfigure}

\title{\LARGE \bf
KINESIS: Motion Imitation for Human Musculoskeletal Locomotion
}

\author{Merkourios Simos$^{1}$,  Alberto Chiappa$^{1}$, Alexander Mathis$^{1}$
\thanks{$^{1}$EPFL}%
\thanks{Contact: {\tt\small alexander.mathis@epfl.ch}}%
\thanks{We thank the reviewers and members of the Mathis Group for helpful feedback. In particular, we thank Bianca Ziliotto, Chengkun Li, Alessandro Marin Vargas, and Johanni Brea for feedback on earlier versions. This project was funded by Swiss SNF grant (310030 212516).}
}

\begin{document}

\maketitle
\thispagestyle{empty}
\pagestyle{empty}

\begin{abstract}

How do humans move? Advances in reinforcement learning (RL) have produced impressive results in capturing human motion using physics-based humanoid control. However, torque-controlled humanoids fail to model key aspects of human motor control such as biomechanical joint constraints and non-linear, overactuated musculotendon control. We present KINESIS, a model-free motion imitation framework that tackles these challenges. KINESIS is trained on 1.8 hours of locomotion data and achieves strong motion imitation performance on unseen trajectories. Through a negative mining approach, KINESIS learns robust locomotion priors that we leverage to deploy the policy on several downstream tasks such as text-to-control, target point reaching, and football penalty kicks. Importantly, KINESIS learns to generate muscle activity patterns that correlate well with human EMG activity. We show that these results scale seamlessly across biomechanical model complexity, demonstrating control of up to 290 muscles. Overall, the physiological plausibility makes KINESIS a promising model for tackling challenging problems in human motor control. Code, videos, and benchmarks are available at https://github.com/amathislab/Kinesis.


\end{abstract}
\section{Introduction}

A hallmark of embodied intelligence is the ability to efficiently navigate the world. In humans and other animals, this capacity to \textit{locomote} translates to a diverse skillset including walking, turning, and running, movements that are driven by hundreds of musculotendon organs, constrained by the intricate joint morphology of the skeletal system~\cite{ramdya2023neuromechanics}. How biological organisms master the complexity of the musculoskeletal system is a crucial question for neuroscience, and it could hold the key to unlocking robust and efficient high-dimensional motor control in robots.

Deep reinforcement learning (DRL) has shown success in the field of physics-based motor control, enabling the control of humanoids that run, jump, and navigate complex terrain~\cite{tessler2024maskedmimic, luo2024universal}. However, due to different scientific goals, these approaches use unrealistic models of the human body, which consist of a simplified kinematic tree and are notably controlled by torque controllers instead of muscles. Realistic musculoskeletal models, which are necessary to capture the true complexity of human locomotion, are receiving increasing attention across scientific domains~\cite{delp2007opensim,loeb2021learning,song2021deep,mathis2024decoding,denayer2025prisma}. Yet, despite the significant progress of the last years~\cite{song2021deep,denayer2025prisma}, current works on muscle-driven locomotion are still limited compared to their torque-controlled counterparts, primarily due to their narrow focus on forward gait~\cite{he2024dynsyn, schumacher2023deprl, schumacher2025natural} or lack of quantitative comparisons to human data~\cite{lee2019scalable, feng2023musclevae, park2025magnet}.

\begin{figure}[t]
    \centering
    \includegraphics[width=1\linewidth]{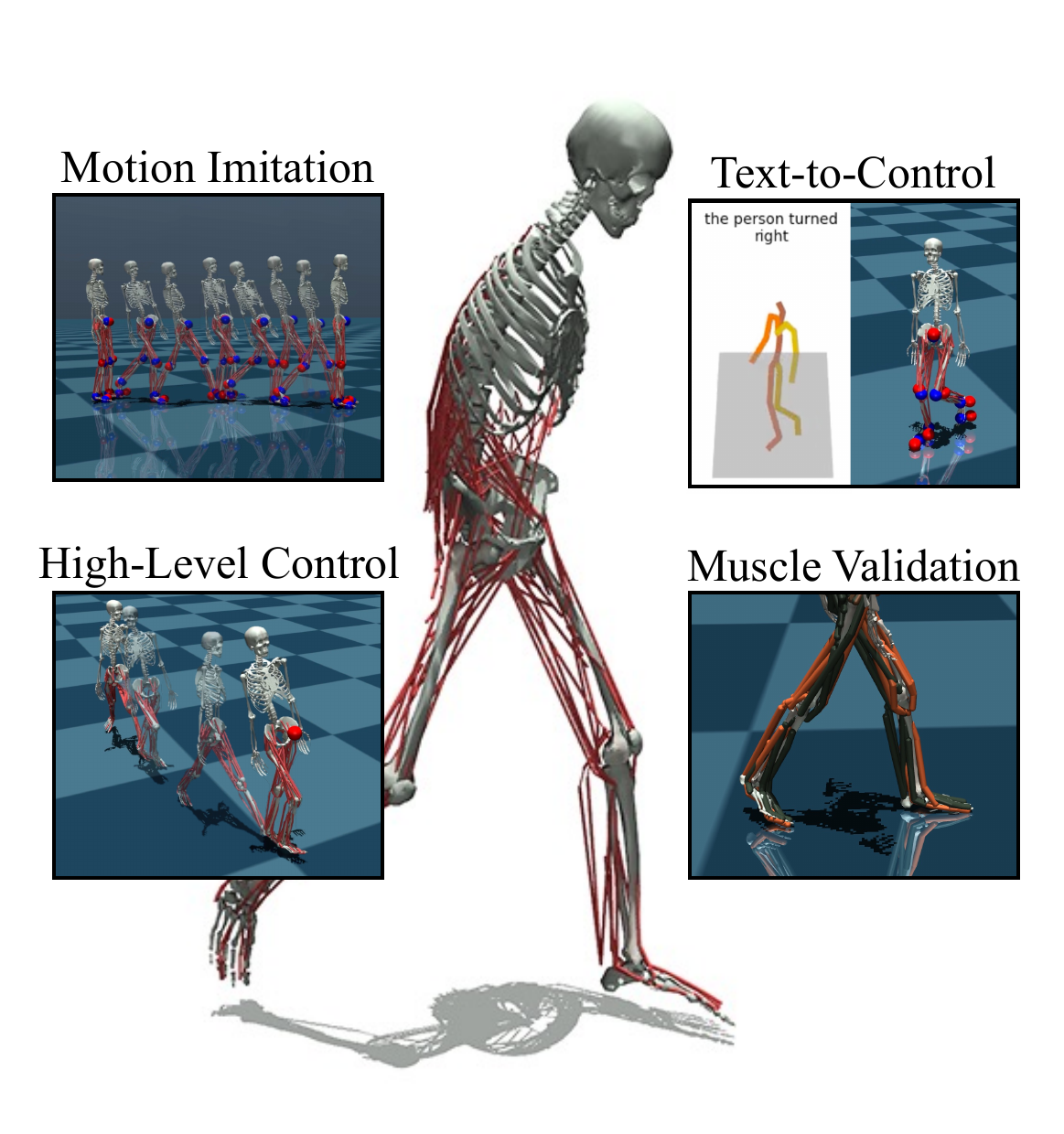}
    \vspace{-30pt}
    \caption{KINESIS is a model-free RL policy for the musculoskeletal control of locomotion. \textbf{Top left:} Our policy is trained on a curated set of MoCap data focusing on locomotion, and can successfully imitate unseen motion clips of the same locomotion types. \textbf{Top right:} KINESIS can be deployed zero-shot on synthetic text-conditioned motion sequences. \textbf{Bottom left:} KINESIS is fine-tuned to reach goal locations, and produces human-like motion without reference motion guidance. \textbf{Bottom right:} Muscle activity patterns produced during motion imitation (active muscles shown in orange, inactive muscles shown in black) correlate well with human electrophysiology data (EMG).}
    \label{fig:abstract-figure}
    \vspace{-14pt}
\end{figure}

We propose \textbf{KINESIS}, a model-free motion imitation framework that facilitates the development of effective and scalable muscle-based control policies of locomotion (Fig.~\ref{fig:abstract-figure}). We trained KINESIS on 1.8 hours of motion capture data that we curated for whole-body locomotion. We implemented this framework on three biomechanically validated musculoskeletal models of increasing complexity, and demonstrate that the learned policies achieve quantifiably high motion tracking performance and high-level controllability through natural language and target-reaching commands. Furthermore, we demonstrate that KINESIS serves as a strong foundation for downstream tasks by fine-tuning the policy to perform football penalty kicks. Finally, we find that the learned policies produce physiologically plausible muscle activity patterns, which we validate by comparisons with human electromyography (EMG) recordings across multiple subjects.

\section{Related Work}

\paragraph{Musculoskeletal Models of the Human Body} Efforts to model the human musculoskeletal system trace back to the Hill-type model, which provided a mathematical formulation of the dynamical properties of muscles~\cite{hill1938heat}, and much later of musculotendon units~\cite{zajac1989muscle}. Since then, many works have incorporated these models into numerical simulations of muscle-based systems~\cite{denayer2025prisma}. In particular, the OpenSim platform provided an open-source tool for the design of accurate musculoskeletal models of the human body~\cite{delp2007opensim, rajagopal2016full}. Unfortunately, it is computationally costly compared to physics engines that are used in robotics, machine learning and RL~\cite{todorov2012mujoco}. To this end, Lee et al. \cite{lee2019scalable} implemented a full musculoskeletal model in DART~\cite{lee2018dart, lee2019scalable}, and incorporated it into a motor skill task. Similarly, Caggiano et al. \cite{caggiano2022myosuite} developed MyoSuite, a suite of physiologically accurate biomechanical models implemented in Mujoco by porting models from OpenSim. Contemporary RL work uses alternative musculoskeletal models with more muscles~\cite{zuo2024self}; however, those are not open-source. In this work, we use three variants of the MyoLeg model from Myosuite, with the most complex variant featuring 290 muscles, more than any other open-source model.

\paragraph{Control of Musculoskeletal Agents} For high-dimensional musculoskeletal control, some works have utilized reference motion data~\cite{lee2019scalable, feng2023musclevae, qin2022muscle, park2025magnet}, whereas others focused on task-oriented learning frameworks \cite{kidzinski2020artificial, caggiano2022myochallenge, park2022generative, schumacher2023deprl,caggiano2023myochallenge,schumacher2025natural, berg2024sar,chiappa2024latent,chiappa2024acquiring,caggiano2024myochallenge,wei2025motion}. MuscleVAE combined motion imitation of various locomotion skills with a muscle fatigue model using a model-based policy~\cite{feng2023musclevae}. More recently, MAGNET introduced a three-stage learning approach for imitation learning of a large MoCap dataset and developed a distilled model for real-time muscle activity generation~\cite{park2025magnet}. While the current lack of available training code for either of these works hinders direct comparison, KINESIS presents several comparative advantages. First, KINESIS is based on a single-stage, model-free algorithm, demonstrating seamless application to three musculoskeletal models with virtually no hyperparameter tuning. Second, we do not rely on model distillation for real-time muscle activity generation; our full model can run at 3.8-times the real speed on the CPU of a commercial laptop. Third, we present comprehensive quantitative metrics for our imitation and EMG comparison results as a baseline for future work.

\paragraph{Comparison with human muscle activity} Electromyography (EMG) signals are a common way to validate the physiological plausibility of biomechanical models fit to motion capture data~\cite{hicks2015my,dewolf2024neuro,schneider2024muscles,denayer2025prisma}. In particular, the simulated muscle activity patterns are verified against physiological sequences measured by EMG. This matching is non-trivial, due to the non-linear, high dimensional, and over-actuated nature of biomechanical systems~\cite{delp2007opensim,loeb2021learning}. Some studies have investigated the correspondence between RL-based muscle control and human muscle activity, but for control policies generating a single motion skill at a time~\cite{qin2022muscle, song2021deep}. MAGNET focused on real-time EMG prediction, but the quantitative correspondence of the generated activity patterns to multi-subject ground truth data remains unclear~\cite{park2025magnet}. Here, we propose a validation benchmark based on EMG data from human subjects during locomotion~\cite{wang2023wearable, scherpereel2023human} that can be applied to any muscle-based control policy modeling leg muscles, and we show that KINESIS significantly outperforms state-of-the-art control policies that use the MyoLeg model~\cite{schumacher2023deprl,he2024dynsyn}.

\section{Methods}

In this section, we describe the three musculoskeletal models from the MyoSuite library~\cite{caggiano2022myosuite}, the curated KIT-Locomotion reference dataset~\cite{AMASS_KIT-CNRS-EKUT-WEIZMANN}, and the motion imitation learning framework.

\vspace{-5pt}

\subsection{Musculoskeletal Model (MyoLeg)}
MyoSuite~\cite{caggiano2022myosuite} is a library built on top of the Mujoco physics simulator~\cite{todorov2012mujoco}, featuring several biologically-accurate models of different human body parts. Among them, MyoLeg is a locomotion-oriented musculoskeletal simulation of the human body that is biologically validated against an OpenSim~\cite{delp2007opensim} model of the full human body by \cite{rajagopal2016full}. In this work, we apply the same training method to three variants of the MyoLeg with increasing complexity:
\begin{itemize}
    \item \textit{Legs}: The simplest version is actuated at the legs by 80 musculotendon units, while the upper body is modeled as a rigid volume with mass.
    \item \textit{Legs+Abs}: This model includes six additional abdominal muscles (86 in total) that enable basic postural control of the upper body.
    \item \textit{Legs+Back}: The most advanced version is a combination of the basic model of the legs with a comprehensive model of the back~\cite{walia2025myoback}, enabling full lumbar control. It consists of 290 musculotendon actuators (Fig.~\ref{fig:abstract-figure}).
\end{itemize}

In all three models, the body dimensions closely match the mean human body shape of the SMPL model~\cite{loper2023smpl}, enabling the seamless use of MoCap datasets.

\vspace{-5pt}

\subsection{Human Motion Dataset}

\paragraph{KIT-Locomotion} We utilize human motion clips from the KIT Motion Dataset~\cite{AMASS_KIT-CNRS-EKUT-WEIZMANN}. KIT is a collection of MoCap datasets under a unified representation framework, containing a variety of full-body motor skills. Because of our focus on the motion of the lower body, we extracted a locomotion-specific subset of 1054 reference motion sequences (KIT-Locomotion), representing 1.8 hours of MoCap data. We selected motion sequences based on their motion description, using key terms such as '\textit{walk}', '\textit{turn}', or '\textit{run}'. We filtered this selection further by evaluating motion categories on feasibility, diversity, and length. In the curated form, {\it KIT-Locomotion} consists of five broad locomotion skills: (1) \textit{walk}, in which the subject walks forward at varying speeds, (2) \textit{gradual turn}, in which
the subject turns to the left or right while walking, (3) \textit{turn in place}, where the subject walks in a straight line, then performs a U-turn and resumes walking, (4) \textit{walk backwards}, where the subject takes backward steps, and (5) \textit{run}, where the subject breaks into a short running stride. We extracted all motions from the AMASS dataset~\cite{mahmood2019amass}, which use the SMPL body format \cite{loper2023smpl}.

\paragraph{Data pre-processing} We downsampled the motions to 30 frames per second before aligning them with the reference frame used in Mujoco. We computed joint rotations in quaternion space, which, along with the global translation of the pelvis, fully describe the motion. For motion imitation, we considered the reference body set \{\textit{head, pelvis, knees, ankles, toes}\}, removing head tracking for the basic model since it lacks upper postural control.

\paragraph{Inverse kinematics} Although motion imitation is performed in Cartesian space, we still need to define the initial position of the model in joint angles, and the joint rotation space of MyoLeg is different compared to the skeletal model that describes motions in the SMPL kinematic structure. Therefore, we performed inverse kinematics to align the model with the reference motion frame. We repeated this process for multiple frames per reference trajectory, thus enabling the initialization at random starting points during the motion.

\subsection{Motion Imitation}
\paragraph{Preliminaries} The motion imitation problem can be cast as a Partially-Observable Markov Decision Process (POMDP) $\mathcal{M} = \langle \mathcal{S}, \mathcal{O}, \mathcal{A}, \mathcal{T}, \mathcal{R}, \gamma \rangle$, characterized by the state space $\mathcal{S}$, the observation function $\mathcal{O}$, the action space $\mathcal{A}$, the transition function $\mathcal{T}$, the reward function $\mathcal{R}$ and the discount factor $\gamma$. At each time step, an agent interacts with the simulated environment in Mujoco to control the MyoLeg model and track a reference motion. Through the observation function $\mathcal{O}: \mathcal{S} \rightarrow \mathbb{R}^{N}$, the environment provides an observation vector $\mathbf{o}_t \in \mathbb{R}^{N}$, describing the proprioceptive state of the body. Specifically, the observation vector includes: the height, the tilt and the velocity of the pelvis, the joint kinematics (position, velocity, angular position and angular velocity for each joint), and the feet contact forces at time $t$, as well as the target reference pose (absolute and relative) at time $t+1$. The agent controls the body by specifying a control signal for each of the $M$ muscles, represented by a vector $\mathbf{a}_t \in \mathcal{A} \subset \mathbb{R}^{M}$.
The next state $\mathbf{s}_{t+1} \in \mathcal{S}$ is determined by the transition function $\mathcal{T}: \mathcal{S} \times \mathcal{A} \rightarrow \mathcal{S}$, defined by the physics simulator and the muscle dynamics. $N$ is equal to 309, 363, and 453 for the basic, medium, and advanced models respectively.

\paragraph{Reward structure} The objective of the agent is to maximize the discounted cumulative reward $R = \sum_{t=0}^{T} \gamma^t r_t$, where $T$ denotes the termination point of the POMDP and $\gamma$ denotes the discount factor. The reward $r_{t} \in \mathbb{R}$ is the output of the function $\mathcal{R}: \mathcal{S}\times \mathcal{A}\rightarrow \mathbb{R}$, which maps the current state $\mathbf{s}_t$ and action $\mathbf{a}_t$ into a reward signal. 
In a motion imitation task, the goal is to reproduce the target movement as faithfully as possible. As in previous motion imitation work~\cite{peng2018deepmimic, luo2023perpetual,luo2024universal}, we defined the imitation component of the reward function as the sum of the negative Euclidean distance between the body's joint position vector $\mathbf{p}_t^i$ and the position vector $\widehat{\mathbf{p}}_{t+1}^i$ of the target pose:

\begin{equation}
r^{pos}_t = \exp\left({-k_{pos} \sum_i||{\mathbf{p}_t^i - \widehat{\mathbf{p}}_{t+1}^i}||^2}\right).
\end{equation}
The joints included in the sum are the pelvis, knees, ankles, toes, as well as the head for the Legs+Abs and L+Back models. We apply an exponential kernel to make the rewards positive and avoid encouraging early termination. The scaling factor $k_{pos}$ controls the rate at which the reward decreases with higher distance from the target motion frame.

Similarly, we defined a velocity reward:

\begin{equation}
r^{vel}_t = \exp\left({-k_{vel} \sum_i||{\mathbf{v}_t^i - \widehat{\mathbf{v}}_{t+1}^i}||^2}\right),
\end{equation}

where $\mathbf{v}^i$ and $\widehat{\mathbf{v}}^i$ denote the joint velocities of the same keypoints for the skeleton and target pose.

Since the musculoskeletal model is overactuated, with antagonist muscles controlling the same joints, there are infinite sequences of muscle activation that produce the same motion. To distinguish between these solutions and discourage simultaneous co-activation of antagonist muscles, we add L1 and L2 regularization in terms of the expended energy at each time step through the reward component:

\begin{equation}
r^{e}_t = - ||\mathbf{m}_t|| - ||\mathbf{m}_t||_2,
\end{equation}

where $\mathbf{m}_t$ is the vector containing the muscle activation signals for time step $t$. Finally, we optionally include an upright reward component:

\begin{equation}
r^{up}_t = \exp\left(-k_{up} \left(\theta_{f,t}^2 + \theta_{s,t}^2\right)\right),
\end{equation}

where $\theta_{f,t}$ and $\theta_{s,t}$ denote the forward and sideways tilt of the pelvis in radians, at time step $t$.

The total reward at each time step is a weighted sum of the reward components, with the individual weights chosen as hyperparameters. 

\begin{figure}
    \centering
    \includegraphics[width=1\linewidth]{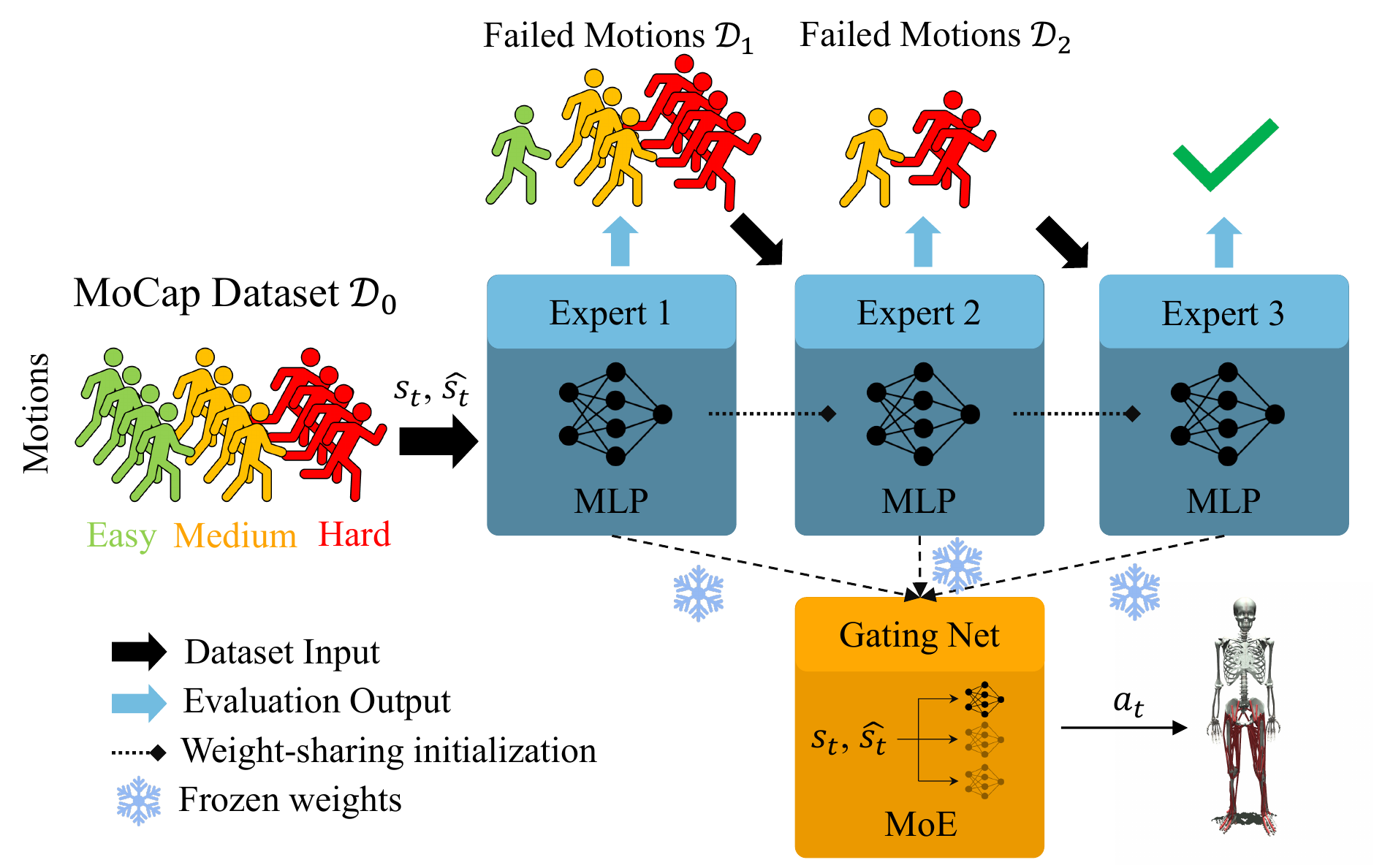}
    \caption{An illustration of the hard negative mining strategy. The first policy network (expert) is trained on the entire dataset, containing five locomotion skills of varying difficulty (here denoted as easy, medium, and hard). After a given number of training epochs, the motions that the first expert successfully imitates are removed from the dataset, and the reduced dataset is used to train a new copy of the expert. The process repeats until the dataset is empty. Finally, a MoE gating network is trained to select the appropriate expert for a specific time step, given the current proprioceptive state and target pose.}
    \label{fig:neg-mining}
    \vspace{-15pt}
\end{figure}

\begin{table*}[t!]
\fontsize{9pt}{9pt}\selectfont
\centering
\caption{Quantitative motion imitation results. Mean and SEM for N=3 evaluation runs.}
\begin{tabular}{@{}lrrrrrrr@{}}
\toprule
                                    & \multicolumn{3}{c}{KIT-Locomotion Train (946 motions)}                      & \multicolumn{3}{c}{KIT-Locomotion Test (108 motions)}  \\ \midrule
\multicolumn{1}{l|}{Model}         & Frame Coverage $\uparrow$ & Success $\uparrow$ & \multicolumn{1}{r|}{ $E_{\mathrm{MPJPE}}$ $\downarrow$} & Frame Coverage $\uparrow$ & Success $\uparrow$ & $E_{\mathrm{MPJPE}}$ $\downarrow$ \\ \midrule

\multicolumn{1}{l|}{Legs-PD}           & 96.98 $\pm$ 0.06 \%       & 90.63 $\pm$ 0.21 \% & \multicolumn{1}{r|}{48.8 $\pm$ 0.16}        & 96.38 $\pm$ 0.79 \%       & 91.36 $\pm$ 1.76 \% & 48.76 $\pm$ 0.08        \\

\multicolumn{1}{l|}{Legs-Direct}   & 98.10 $\pm$ 0.05 \%       & 93.27 $\pm$ 0.26 \% & \multicolumn{1}{r|}{42.72 $\pm$ 0.22}        & 97.99 $\pm$ 0.35 \%       & 94.14 $\pm$ 0.91 \% & 43.04 $\pm$ 0.42       \\

\multicolumn{1}{l|}{Legs+Abs}  & 98.99 $\pm$ 0.06 \%       & 96.48 $\pm$ 0.12 \% & \multicolumn{1}{r|}{44.01 $\pm$ 0.22}        & 98.63 $\pm$ 0.41 \%       & 95.68 $\pm$ 1.10 \% & 44.17 $\pm$ 0.61        \\

\multicolumn{1}{l|}{Legs+Back} & \textbf{99.22} $\pm$ \textbf{0.06} \%       & \textbf{97.04} $\pm$ \textbf{0.38} \% & \multicolumn{1}{r|}{\textbf{42.26} $\pm$ \textbf{0.04}}        & \textbf{99.43} $\pm$ \textbf{0.03} \%         & \textbf{96.91} $\pm$ \textbf{0.50} \%   & \textbf{42.84} $\pm$ \textbf{0.29}        \\
\bottomrule
\end{tabular}
\label{tab:results}
\vspace{-10pt}
\end{table*}

\paragraph{Muscle actuation} We tested two widely-used motor control schemes to map the policy output into the activation signal. First, we followed an indirect control scheme, in which the policy output $\mathbf{a}_t \in \mathbb{R}^{M}$ represents the desired next-step muscle lengths~\cite{luo2023perpetual, feng2023musclevae, lee2019scalable}. A proportional-derivative (PD) controller is responsible for finding the force which a muscle should apply for the joint to reach the desired position, according to the equation $F_{i} = F_i^{\mathrm{peak}} \left( k_p (a_i - l_i) - k_dv_i \right)$, where $i$ denotes the muscle index, $F^{\mathrm{peak}}$ denotes the maximum force that the muscle can exert, $l$ denotes the current muscle length, $v$ denotes current muscle velocity, while $k_p$ and $k_d$ are scaling factors that are empirically determined and remain fixed.

To convert muscle force into muscle activity $\mathbf{m}_t \in \mathrm{R}^{M}$, we use the force-length-velocity relationship:

\begin{equation}
    m_i = \frac{F_{i} - F_P(l_i)} {F_L(l_i) F_V(v_i)},
\end{equation}

where $F_P$ denotes the force exerted by the muscle's passive element as a function of its length, $F_L$ denotes the active isometric force exerted by the muscle as a function of its length, and $F_V$ denotes the active force exerted by a maximally activated muscle at its resting length, as a function of its velocity. These parameters are provided in MyoLeg.

Alternatively, we implemented a direct mapping $\mathbf{m}_t = (\mathbf{a}_t + 1) / 2$, which simply rescales the policy output into the muscle activation range. We compared these two methods in the Legs model, and we found that the direct control scheme reduced training and simulation time significantly, while achieving better results. Therefore, we used the direct control scheme for the Legs+Abs and Legs+Back models, as well as for the downstream tasks.

\begin{figure*}
    \centering
    \includegraphics[width=0.95\linewidth]{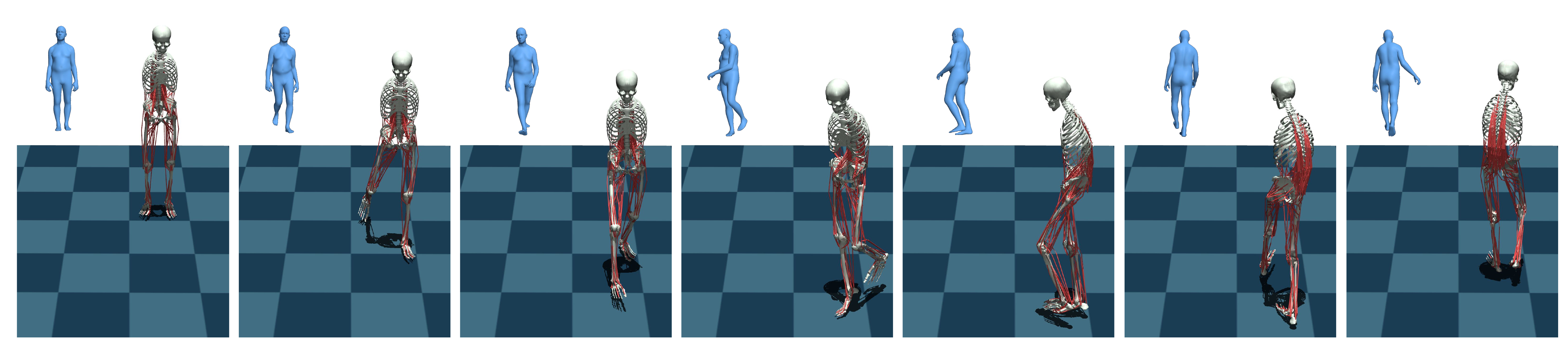}
    \vspace{-8pt}
    \caption{The Legs+Back musculoskeletal model is imitating a ``\textit{Turn in Place}" motion; a rendering of the reference motion is shown in blue.}
    \label{fig:kit-example}
    \vspace{-15pt}
\end{figure*}


\subsection{Reinforcement Learning}

In RL, the agent interacting with the environment is defined by the policy $\pi(\mathbf{a}_t | \mathbf{o}_t)$, which specifies the conditional distribution over actions $a_t \in \mathbb{R}^{M}$ given an observation $\mathbf{o}_t \in \mathbb{R}^{N}$. We built on the motion imitation approach by Luo et al.~\cite{luo2023perpetual}, using a training set $\mathcal{D}$ with reference motion sequences, and following a Mixture-of-Experts (MoE) approach. 

Training episodes begin by sampling a random reference motion from $\mathcal{D}$ and a random starting frame, setting the musculoskeletal model to the reference state, and initiating environment dynamics. The distance between the reference frame and the model is computed as the episode progresses. If any tracked keypoint deviates more than 0.15 meters from its reference, the episode ends early; otherwise, it concludes at the end of the sequence without further reward, and a new motion is sampled.

We modeled each expert policy with a 6-hidden layer [2048, 1536, 1024, 1024, 512, 512] multilayer perceptron (MLP)~\cite{luo2023perpetual}. The control policy $\pi_\theta(\mathbf{a}_t | \mathbf{o}_t) = \mathcal{N}(\mu(\mathbf{o}_t, \mathbf{\theta}), \Sigma)$ is a multivariate Gaussian distribution with a covariance matrix $\Sigma$ and mean $\mu(\mathbf{o}_t, \theta)$, where $\theta$ represents the learned policy network parameters. The policy is trained using proximal policy optimization (PPO) and Lattice, an exploration method that is adapted for high-dimensional control~\cite{schulman2017proximal, chiappa2024latent}. The same MLP architecture models the value function.

\paragraph{Hard negative mining } The agent must generate a diverse array of control plans to successfully imitate the reference motions. However, the policy required for one motion sequence may be incompatible with policies learned for other motions. To address this, we followed~\cite{luo2023perpetual} and we employed an ensemble policy based on negative hard mining (Fig.~\ref{fig:neg-mining}). In more detail, the training phase began by sampling motions from the entire training dataset. When the  imitation success rate reached a plateau, we identified and isolated the motion sequences that the policy failed on. We then initialized a copy of the policy with the currently-learned weights, and we initiated a new round of training using only the isolated motion sequences. We continued this process until there were no failed motion sequences left.

\paragraph{Mixture of experts} Ultimately, we obtained a set of different experts that covered the entire training dataset. To combine these policy networks into a single, universal policy, we froze their parameters and trained an MoE model on the entire training set~\cite{jacobs1991adaptive}. The gating network is a three-layer [1024, 512, 256] MLP that takes as input the current observable state and outputs a vector $\widehat{\mathbf{a}}_t \in \mathbb{R}^n$ that determines which expert's output will be used to actuate the model. We used a softmax transformation so that $\widehat{a}_t^i$ denotes the probability of using the $i$'th expert's output as the selected action. Essentially, the MoE network acts as a gating mechanism that stochastically selects an expert based on the current observations (Fig.~\ref{fig:neg-mining}). We found that three experts were sufficient to achieve high imitation performance for all three musculoskeletal models.

\begin{table}
\centering
\caption{KINESIS Hyperparameters. Different values for individual models are denoted as Legs/Legs+Abs/Legs+Back}
\begin{tabular}{|l|l|l|l|}
\hline
\textbf{Hyperparameter} & \textbf{Value} & \textbf{Hyperparameter} & \textbf{Value} \\ \hline
Sim. timestep & $150^{-1}$ & Reward $w_{vel}$ & 0.2 \\ \hline
Control interval & 5 & Reward $w_{e}$ & 0.1/0.2/0.2 \\ \hline
PD controller $k_p$ & 1 & Reward $w_{up}$ & 0.1/0.0/0.0 \\ \hline
PD controller $k_d$ & 1 & Reward $w_{tracking}$ & 0.8 \\ \hline
Reward $k_{pos}$ & 200 & Reward $w_{success}$ & 20 \\ \hline
Reward $k_{vel}$ & 5 & PPO $\gamma$ & 0.99 \\ \hline
Reward $k_{up}$ & 3 & PPO $\epsilon$ & 0.2 \\ \hline
Reward $w_{pos}$ & 0.6 & PPO $\tau$ & 0.95 \\ \hline
PPO learning rate & $5 \times 10^{-5}$ & PPO optimizer & Adam \\ \hline
Activation function & SiLU & & \\ \hline
\end{tabular}
\label{tab:kinesis-params}
\vspace{-18pt}
\end{table}

\subsection{High-level Control Tasks}

Beyond motion imitation, a truly intelligent agent should be able to apply its learned skills to novel, high-level control problems. We selected three downstream tasks—text-to-control, target goal reaching, and football penalty kick—that progressively test the policy's ability to generalize its learned locomotion priors to increasingly abstract and interactive challenges.

\paragraph{Text-to-control} Natural language is an ideal medium to convey high-level behavioral goals. To this end, we leveraged Human Motion Diffusion Model (MDM) to transform the motion imitation task into a text-conditioned motor control task. MDM generates motion sequences as keypoints that follow the SMPL model~\cite{tevet2023human}, which is perfectly aligned with the dimensions of our model's skeleton. We extracted the kinematics from the generated motions and initiated the motion imitation task without additional training, evaluating the musculoskeletal model in a zero-shot manner.

\paragraph{Target goal reaching} Although the abstraction of natural language is powerful, sometimes a more concrete goal formulation is favorable; for example, when we need to specify an exact location to locomote to. To test whether the locomotion skills acquired by KINESIS during motion imitation training can be transferred to other tasks without relying on reference data, we designed a target goal-reaching task. The objective was for the model to reach a designated position within a 2$\times$2 m range from a given initial pose and facing direction. We provided the policy with proprioceptive input and target keypoints, but we set all ``targets" to the current position of the respective body part, apart from the target root position, which was set to the target goal, following~\cite{luo2023perpetual}. We updated the reward function to better match the high-level task by replacing the position and tracking velocity rewards with two new components, $r_t^{track} = \delta_{t-1} - \delta_t$, and $r_t^{success} = \mathbf{1}_{\{\delta_t < 0.1\}}$, where $\delta_t$ denotes the distance between the root and the target position at time $t$~\cite{won2022physics}. We set the duration of each episode to 5 seconds, and terminated the episode early if the height of the root dropped below 0.7 m. To leverage the locomotion priors of KINESIS, we used the second expert policy as the starting point for fine-tuning.

\paragraph{Penalty kick} Locomotion is often combined with other goal-directed movements that involve interactions with objects, such as a football. To test the capacity of KINESIS in this challenging context, we deployed the Legs+Back goal-reaching policy on a penalty-kick task, where the model must step to the ball and kick it towards the net while beating the goalkeeper. Similar to the goal reaching task, the observation space remains the same, and the position of the ball is given to the policy as the target root position. To incentivize the policy to score goals, we simply provide the velocity of the ball towards the net as large positive reward. The environment was adapted from the 2025 MyoChallenge~\cite{MyoChallenge2025}.

\begin{figure*}[t!]
    \centering
    \includegraphics[width=1\linewidth]{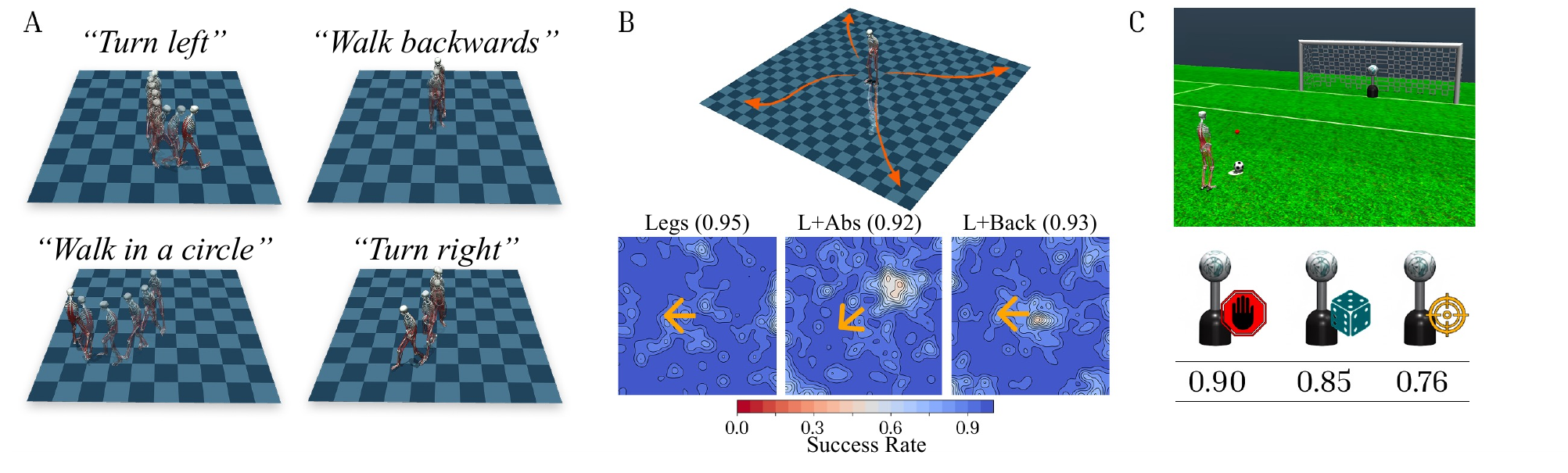}
    \vspace{-20pt}
    \caption{Examples of musculoskeletal motor control during high-level tasks. \textbf{A}: \textit{Text to motion}. Four examples of natural language prompts that KINESIS executes zero-shot. \textbf{B}: \textit{Target goal reaching}. The agent is tasked with reaching random positions in $\mathbb{R}^2$ space starting from an initial position and orientation (orange arrows); success rate heatmaps are shown for the three musculoskeletal models (average success rate in parentheses). \textbf{C}: \textit{Penalty kick}. The agent is tasked with beating the goalkeeper in a penalty kick, starting from a random position behind the ball. The policy consistently beats three goalkeeper strategies of increasing difficulty: a static strategy (red), a random walk strategy (blue), and an active blocking strategy (yellow); success rate after 100 kicks.}
    \label{fig:downstream}
    \vspace{-18pt}
\end{figure*}

\section{Experiments}

\subsection{Implementation Details}

\paragraph{Training and inference} All models were trained on a single NVIDIA A100 GPU and 128 CPU threads. We used 128 parallel environment instances. Each training step consisted of experience accumulation for 51\,200 steps, gathered in total from all environments, followed by 10 rounds of policy optimization using PPO~\cite{schulman2017proximal}. All hyperparameters are shown in Table~\ref{tab:kinesis-params}.

Training was performed at a simulation frequency of 150 Hz and a control frequency of 30~Hz. At those settings, inference on the CPU of a modern laptop is 3.8 times faster than real-time. The simulation and control frequencies can be increased significantly, without loss of performance. 

\paragraph{Datasets} We trained KINESIS on the training split of the curated KIT-Locomotion dataset, consisting of 946 motions and five locomotion skills (walking, walking while turning, turning around in place, walking backwards, running). We evaluated the learned policy on the test split, containing 108 unseen motions from the same skill categories. For text-to-motion validation, we used MDM to generate synthetic motion trajectories for each skill category. Text prompts were chosen to ensure that generated motions matched the learned skills in style and velocity. The datasets were the same across all three musculoskeletal models.

\paragraph{Evaluation Metrics} We report the motion imitation frame coverage as the percentage of the dataset (in frames) that the model successfully imitates. We also report the trial success rate, considering a trial successful when the tracked body markers of the model remain within $0.5$ meters of the reference motion on average, throughout the trial trajectory. Furthermore, we report the global mean per-joint position error (MPJPE) in millimeters, averaged across all frames. All motion imitation experiments were performed three times. In the target goal-reaching task, we simulated 1000 random trials, and considered successful those trials that ended with the model within $0.1$ meters of the target. In the penalty kick task, we simulated 1000 kicks with the initial position chosen randomly behind the ball, and considered successful the penalty kicks where a goal was scored against the goalkeeper while the model remained standing.


\vspace{-8pt}

\subsection{Results: Motion Imitation}

\paragraph{Performance on KIT-locomotion Dataset.} A control policy trained using negative mining and three motor experts combined with a MoE module achieved high performance on all locomotion skills (Table~\ref{tab:results}; example in Fig.~\ref{fig:kit-example}). Imitation performance was higher with direct control vs. PD-control (Legs-PD vs. Legs-Direct); model training and inference was also faster. We thus focus on direct control from now on. The two more complex models (Legs+Abs, Legs+Back) achieved slightly higher imitation performance, highlighting the importance of the upper body for locomotion and the good scaling properties of our method.

\vspace{-6pt}

\subsection{Results: High-level Control Tasks}

\paragraph{Text-to-Control} As a first task, we considered text-to-motion control, which we evaluated qualitatively. We prompted MDM with textual descriptions of simple locomotion (e.g.,``the person walked in a circle"). KINESIS could successfully imitate the generated motions in a zero-shot manner, across all three musculoskeletal models (see example trajectories for the Legs+Back model in Fig.~\ref{fig:downstream}A). 

\paragraph{Target Goal Reaching} The fine-tuned policy achieved strong performance on the target reaching task (Fig.~\ref{fig:downstream}B). KINESIS consistently navigated to the target and remained there for the duration of the episode. Notably, the trained agents utilized not only the locomotion skills acquired during the original imitation learning, but also additional skills that had not been previously encountered, such as side steps and diagonal backward motion. Only the Legs+Abs model struggled somewhat for targets right behind the initialization (success rate around 60\%). Importantly, the agents retained a realistic gait after fine-tuning on this task. 
We highlight the importance of the imitation learning phase in achieving natural gait in the Appendix.

\paragraph{Football penalty kick} The Legs+Back model is placed in a penalty kick task against a simulated goalkeeper that follows three adversarial strategies of increasing difficulty. More specifically, the goalkeeper can either remain static at the center of the goal, move randomly under Brownian noise, or track the lateral position of the ball with a bounded velocity. We found that adding a simple reward component for ball velocity enabled the fine-tuned KINESIS policy to consistently beat the goalkeeper~(Fig.~\ref{fig:downstream}C). 

\begin{figure}
    \centering
    \includegraphics[width=0.9\linewidth]{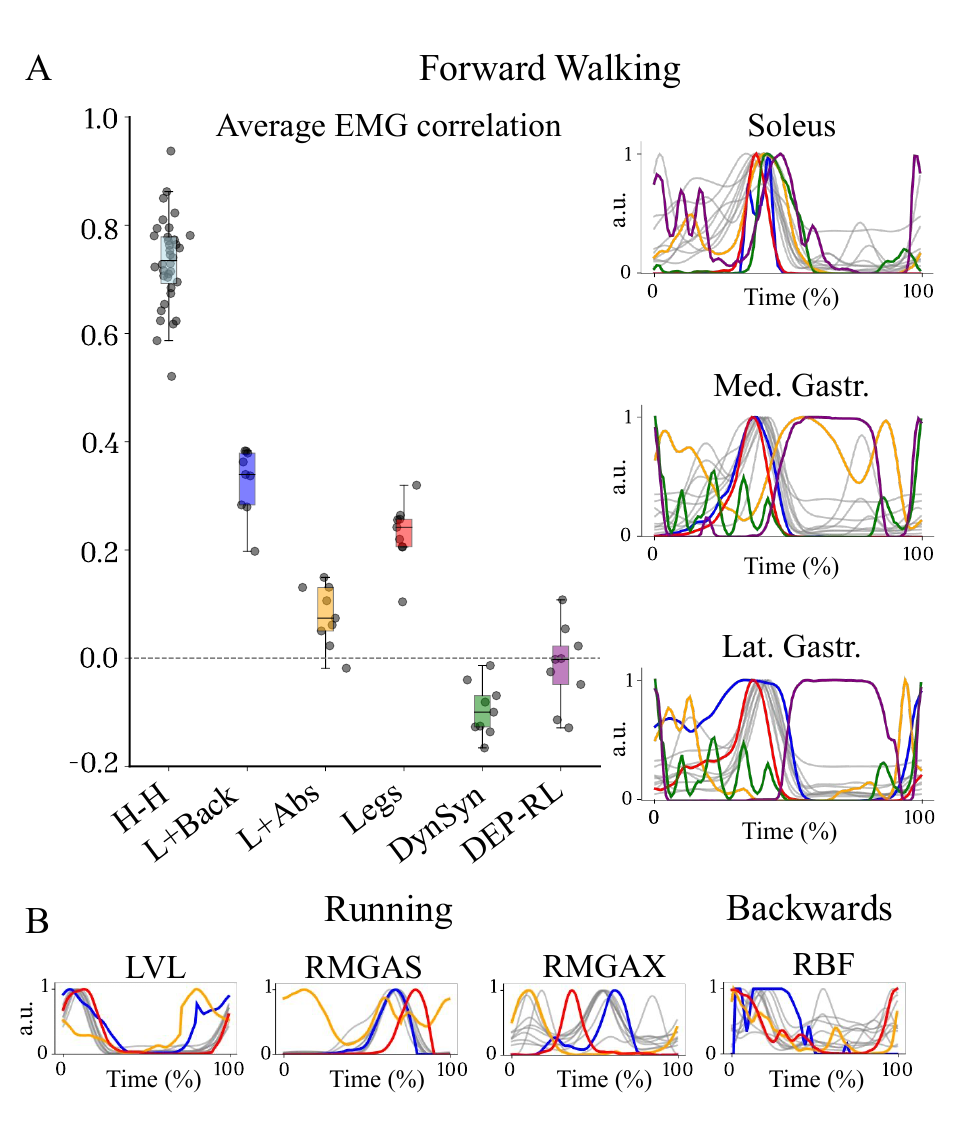}
    \caption{Correlation of generated gait-averaged muscle activity patterns with recorded human EMG data. \textbf{Left, forward walking:} When averaging over gait cycles and muscles, the correlation of muscle activity patterns produced by KINESIS with EMG data, across embodiments, is significantly higher compared to alternative methods using the same musculoskeletal model. Between different KINESIS models, the Legs+Back model has the highest correlation. Individual points denote the average muscle activity correlation between the model and one subject ($N=9$), and pairwise correlation between human subjects (H-H, $N=36$). \textbf{Right}: Example gait-averaged muscle activity patterns during forward walking. \textbf{Bottom}: Example gait-averaged muscle activity patterns during running (LVL: left vastus lateralis, RMGAS: right medial gastrocnemius, RGMAX: right gluteus maximus) and backward walking (RBF: right biceps femoris).}
    \label{fig:muscle-corr}
    \vspace{-20pt}
\end{figure}

\vspace{-4pt}

\subsection{Results: EMG Analysis}

Imitation performance and downstream controllability are great challenges to test the capacity of models for human motion. We hypothesized that pretraining with motion imitation also leads to better alignment with muscle patterns. To assess how closely muscle-based control policies mimic biological muscle control, we developed a benchmark that measures the correlation of generated muscle activity with recorded human electromyography (EMG) measurements. For this evaluation, we adapted two datasets that include recordings from leg muscles in different subjects during various forms of locomotion~\cite{wang2023wearable, scherpereel2023human}. We compared the musculoskeletal models trained with KINESIS with two recent locomotion baselines, DynSyn and DEP-RL~\cite{he2024dynsyn, schumacher2023deprl}, that perform forward locomotion using the Legs model.

First, we quantified the inherent variability across human subjects by averaging the individual muscle activity patterns across the periodic gait cycle~\cite{wang2023wearable}, and calculating the pair-wise Pearson correlation~(Fig. \ref{fig:muscle-corr}A). Human-to-human variability represents a strong upper bound for model performance, since a model within that range would be just as good as another human's muscle dynamics for predicting muscle activity. Computing the correlation of generated average activity patterns across motion types and musculoskeletal models, we found that KINESIS achieves positive correlation and significantly outperforms both baselines in the forward locomotion test. KINESIS was able to capture the qualitative characteristics of the muscle activity patterns; the muscles shown in Fig.~\ref{fig:muscle-corr}A are especially relevant for identifying gait disorders~\cite{lenhart2014empirical}. In addition, KINESIS matches the human EMG patterns for several other muscles during running and backward walking (Fig.~\ref{fig:muscle-corr}B).

Interestingly, KINESIS Legs+Abs produced less human-like muscle activity compared to Legs, despite the added postural control afforded by the abdominal muscles. Our collaborators later found an unrealistic increase in upper body weight in this particular embodiment. This discrepancy might explain the drop in EMG correlation and highlights the value of this analysis in evaluating both musculoskeletal models and policies.

\begin{figure}[t!]
    \centering
    \includegraphics[width=0.8\linewidth]{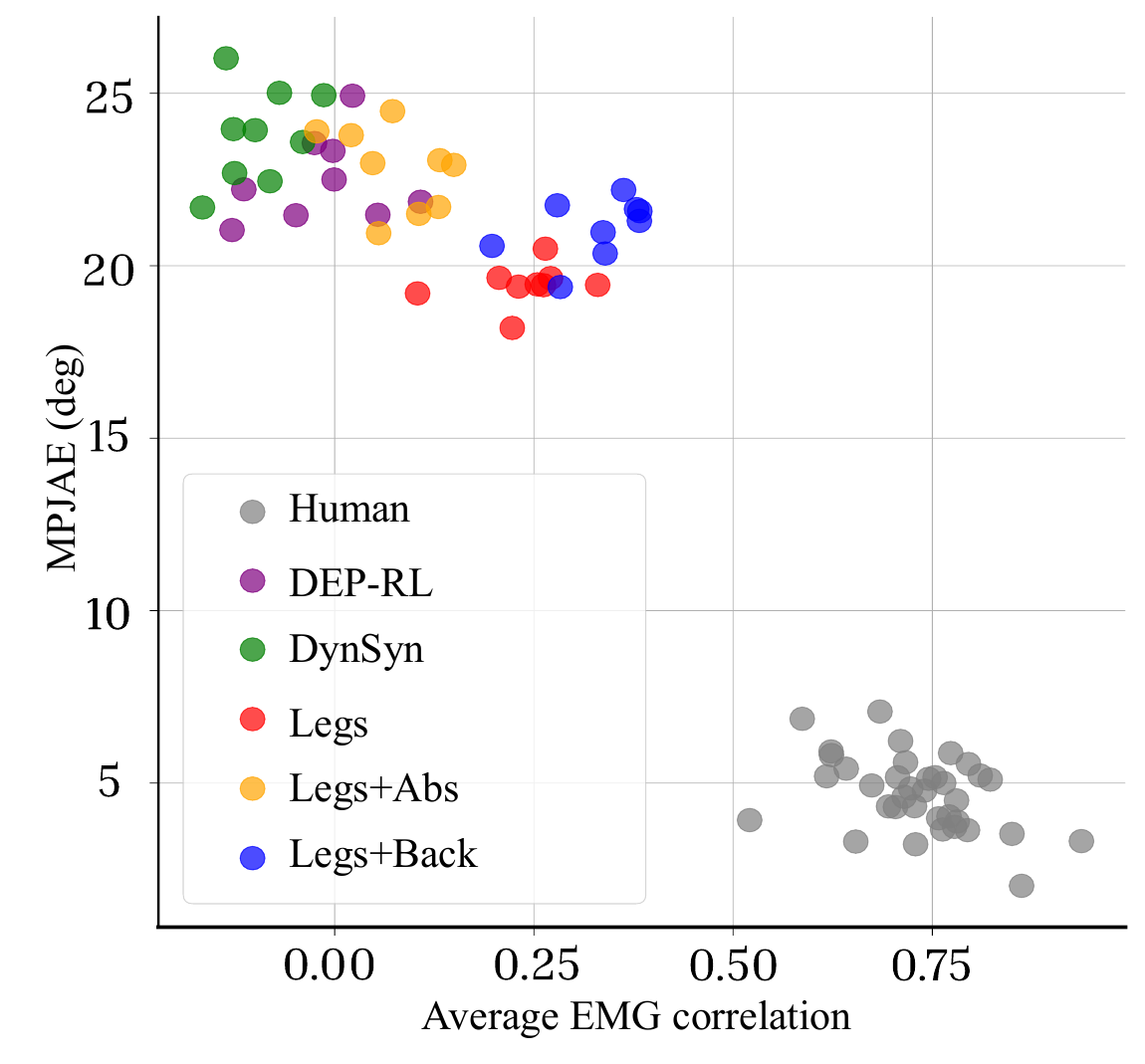}
    \vspace{-6pt}
    \caption{Average EMG correlation between models and human data versus mean joint angle error (MPJAE) between the movement generated by the model and the human trajectory.}
    \label{fig:muscle-joint-corr}
    \vspace{-18pt}
\end{figure}

RL from scratch often does not necessarily produce biologically plausible behavior (e.g.,~\cite{schumacher2025natural,chiappa2024acquiring}), whereas imitation learning explicitly ensures alignment with human motion. We therefore evaluated whether the advantage of KINESIS in terms of EMG correlation was strictly due to this gait similarity (Fig.~\ref{fig:muscle-joint-corr}). We calculated the mean joint angle error (MPJAE) between generated and human gaits by using the available kinematic data, considering four joints of the left leg that showed high correlation across human subjects (hip flexion, hip adduction, knee angle, ankle angle). As expected, we found that DynSyn and DEP-RL had high MPJAE. However, the MPJAE of KINESIS was also significantly higher compared to inter-subject values. Despite achieving only minor improvements in kinematic fidelity over alternative methods, KINESIS learned substantially more accurate EMG patterns. We believe this is a promising result in support of our initial hypothesis, and suggests that motion imitation methods can close the gap to the human ceiling further as future musculoskeletal models evolve.

\vspace{-5pt}

\section{Conclusion}


We presented KINESIS, an RL-based motion imitation framework for musculoskeletal locomotion that demonstrates high imitation performance on 1.8h of MoCap data and effective deployment across downstream tasks. To our knowledge, KINESIS is the first model-free policy combining diverse locomotion skills while generating human-like EMG patterns without being explicitly trained to do so~\cite{denayer2025prisma}. 
Later KINESIS also formed the basis of our winning solution to Myochallenge 2025~\cite{MyoChallenge2025}. More importantly, it could serve as an \textit{in silico} model of the human sensorimotor system to tackle fundamental questions about motor control; a demonstration of this potential in the context of muscle synergies is elaborated in the Appendix.
Significant challenges remain towards a better understanding of human sensorimotor control, including more comprehensive training datasets, and refined musculoskeletal models --- nevertheless, our results suggest a promising path.










\clearpage

{
    \small
    \bibliographystyle{IEEEtran}
    \bibliography{main}
}

\section*{APPENDIX}

\subsection*{Motion Imitation: Data Processing}

To ensure compatibility with the skeletal models, the KIT Dataset is extracted from the AMASS motion capture (MoCap) archive~\cite{mahmood2019amass, mandery2015kit}. Each motion consists of a textual description, video and body shape metadata, and motion data in numerical array form. They key motion data necessary to process the motion are the ``poses" data, which contain the SMPL-H pose in axis-angle space~\cite{loper2023smpl}, and the ``trans" data, which contain the global root translation at each frame.

We follow Luo et al.~\cite{luo2023perpetual} to align the KIT motions to the correct skeletal reference frame~\cite{luo2023perpetual}. First, we perform downsampling to reduce the MoCap frame rate from 100 to 30 frames per second (FPS). Then, we map the SMPL body joints to 24 joints of a template skeleton model in Mujoco, which shares the same dimensions as the MyoLeg musculoskeletal model. We transform the joint rotations from their axis-angle representation to quaternions. Following that, we implement these rotations on the template skeleton model, and compute the global rotation of each joint. Finally, we rotate the skeleton such that the z-axis vector is pointing upwards, and extract the transformed local joint rotations.

\subsection*{Text-to-motion}

We use the Human Motion Diffusion Model (MDM) by Tevet et al.~\cite{tevet2023human} to generate text-conditioned motion sequences resembling the fundamental locomotion skills that KINESIS learns during motion imitation. In particular, we use five textual prompts and generate 10 samples for each one, totaling 50 motions: 
\begin{itemize}
\item ``the person walked forward", 
\item ``the person turned left", 
\item ``the person turned right", 
\item ``the person walked in a circle slowly", 
\item ``the person walked backward slowly". 
\end{itemize}

Each motion has a duration of three seconds. We encourage readers to refer to the official implementation page for demonstrations.


\subsection*{High-Level Control}

When fine-tuning KINESIS for high-level control, we keep the same structure for the observations and the policy outputs. More specifically, we provide the policy with proprioceptive input and target keypoints, but we set all ``targets" to the current position of the respective body part, apart from the target root position, which is set to the target position, similar to Luo et al.~\cite{luo2023perpetual}. During the fine-tuning phase, the musculoskeletal model is initialized at an upright position at the center of a 2-by-2 $m^2$ arena. The agent is tasked with reaching a target root position sampled uniformly at random within the arena. We update the reward function to better match the high-level task by replacing the position and tracking velocity rewards with two new components:

\begin{align}
    r_t^{track} &= \delta_{t-1} - \delta_t, \\
    r_t^{success} &= \mathbbm{1}_{\{\delta_t < 0.1\}},
\end{align}

where $\delta_t$ denotes the distance between the root and the target position at time $t$, which is inspired  by Won et al.~\cite{won2022physics}. We set the duration of each episode to 150 frames (5 seconds), and terminate the episode early if the height of the root drops below 0.7 meters, indicating that the musculoskeletal model has lost its balance.

To reach a high performance in the target goal reaching task, we used the second expert policy from KINESIS and trained on the new task for 5'000-10'000 additional epochs, depending on the embodiment. We also trained a baseline policy on the Legs embodiment, which used the same architecture as the KINESIS experts, on the target reaching task for 5'000 epochs, but skipping the motion imitation training. While the baseline model learns to stand upright and moves towards the target, it employs an unrealistic motion strategy in which one leg is stiff and on the ground, and the other leg is used to propel the model forward (Figure S\ref{fig:joint_analysis}). These results suggest that the imitation learning phase is crucial for the acquisition of natural and robust locomotion priors.

\begin{suppfigure}
    \centering
    \includegraphics[width=1\linewidth]{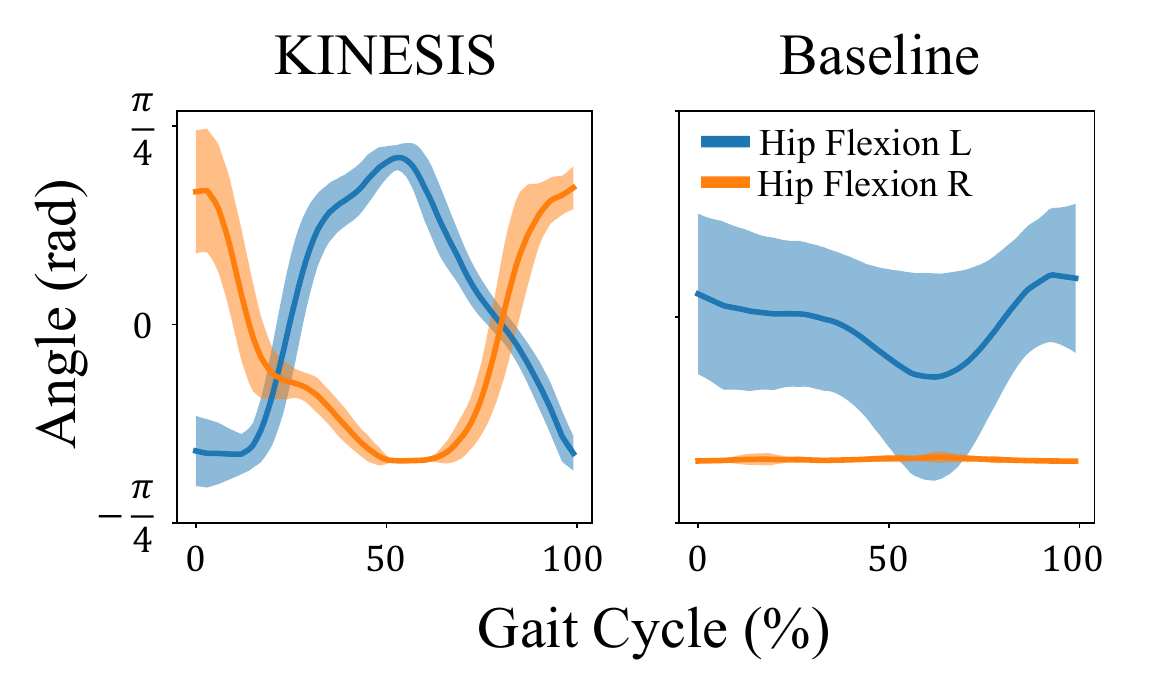}
    \caption{Joint angle analysis: KINESIS produces human-like, symmetric motion where both legs move in a cyclical fashion, as indicated by the average hip flexion along the gait cycle. In contrast, the baseline policy produces unrealistic motion where the right leg is still, while the left leg ``drags" the body forward. Shaded areas indicate standard deviation. Data were recorded during forward motion over a period of 30 seconds for each policy.}
    \label{fig:joint_analysis}
\end{suppfigure}

\subsection*{Motor Synergies for Locomotion}

\begin{suppfigure}
    \centering
    \includegraphics[width=0.9\linewidth]{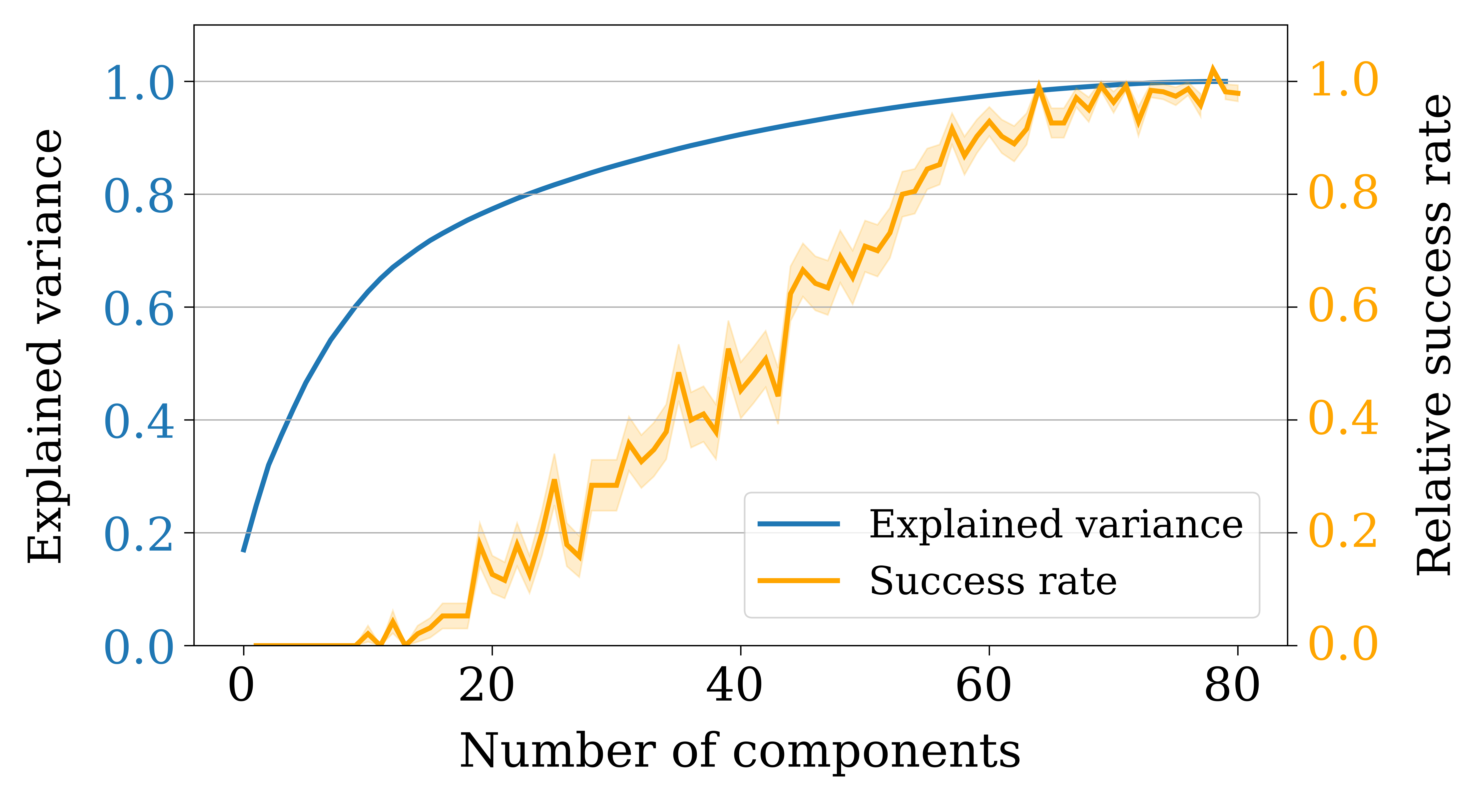}
    \caption{Comparison of two different methods for assessing control signal dimensionality. Dimensionality reduction (blue) overestimates the role of early principal components, which are insufficient to carry out the motor task (orange). The orange shaded region represents the 95\% confidence interval based on 300 trials.}
    \label{fig:csi-ev}
\end{suppfigure}

A fundamental question in motor control is how the human nervous system overcomes and potentially benefits from the over-actuated and high-dimensional nature of the musculoskeletal system (known as the ``Bernstein problem"~\cite{bernstein1967coordination,loeb2021learning}). 

Previous studies have shown that a low number of components, commonly referred to as ``muscle synergies", are sufficient to account for a large proportion of the explained variance (EV) of EMG signals  during motor tasks~\cite{d2003combinations,tresch2006matrix,tresch2009case}. These results have led, according to some researchers, to the influential hypothesis that the brain ``copes" with the complexity of motor control by actually only controlling a few synergies; this idea has also been formalized to create ``brain-inspired" control policies of humanoid characters~\cite{ficuciello2019vision,he2024dynsyn,berg2024sar}. However, techniques that focus solely on explaining the variance of the motor control ignore the possibility that small errors in the muscle control signal could lead to catastrophic failure in terms of task performance~\cite{loeb2021learning}. Therefore, instead of the common signal reconstruction metric, it is much more informative to quantify the dimensionality of a control signal by the number of components that are required to retrieve full task performance~\cite{chiappa2024acquiring}. 

To this end, we recorded the muscle activity patterns generated by KINESIS-Legs during the target goal reaching task, and we decomposed it using Principal Component Analysis. Like in many previous studies, we observed that a relatively low number of components was sufficient to explain a significant proportion of the signal's EV~(Figure S\ref{fig:csi-ev}). We then repeated the task, this time projecting the control signal onto a subspace of principal components before providing it to the model (see Chiappa et al. \cite{chiappa2024acquiring} for the inspiration for this method used for dexterous hand use). We found that the task performance was only recovered when using a much higher number of components. This highlights that for task success, almost all dimensions are required~(Figure S\ref{fig:csi-ev}). Overall, these results show that embodied musculoskeletal policies, such as KINESIS, can provide a powerful alternative to traditional methods of investigating human motor control.

\end{document}